\documentclass[table]{gtech}
\PassOptionsToPackage{table, usenames, dvipsnames}{xcolor}
\usepackage{xcolor}
\usepackage{amssymb}
\usepackage{multirow}
\usepackage{bigdelim}
\usepackage{longtable}
\usepackage{tabularray}
\usepackage{wrapfig}
\usepackage[most]{tcolorbox}
\usepackage{url}
\usepackage{float}
\usepackage{datatool}
\usepackage{enumitem}
\usepackage{subcaption} %
\usepackage[justification=centering]{caption}

\RequirePackage{tgpagella} %
\RequirePackage{mathpazo}  %
\RequirePackage{inconsolata} %
\usepackage{makecell}

\usepackage{booktabs} %
\usepackage{adjustbox}
\usepackage{pgfplots}
\pgfplotsset{compat=1.18}
\usepackage[utf8]{inputenc} %
\usepackage[T1]{fontenc}    %
\usepackage{hyperref}       %
\usepackage{url}            %
\usepackage{booktabs}       %
\usepackage{amsfonts}       %
\usepackage{nicefrac}       %
\usepackage{microtype}      %
\usepackage{xcolor}         %
\usepackage{xspace}
\usepackage{amsthm}   %
\usepackage{amsmath}  %
\usepackage{amssymb}  %
\usepackage{bm}       %
\usepackage{algorithm}
\usepackage{algpseudocode}

\theoremstyle{definition}

\renewcommand{\title}[1]{\newcommand{\titlelist}{{\huge\fontfamily{optimistic}\selectfont #1}}}

\newcommand{\ignore}[1]{}

\usepackage{arydshln}
\definecolor{CQColor}{rgb}{0.0,0.0,1.0} %

\usepackage{colortbl}
\usepackage{amssymb}
\usepackage{pifont}
\usepackage{booktabs,multirow}
\usepackage{makecell}
\usepackage{tabulary}
\usepackage{fontawesome5}
\usepackage{bbding}
\usepackage{multicol}

\newlength\savewidth

\title{How Do Decoder-Only LLMs Perceive Users? Rethinking Attention Masking for User Representation Learning}

\author[1,2\dag]{Jiahao Yuan}
\author[1\dag]{Yike Xu}
\author[1]{Jinyong Wen}
\author[1,*]{Baokun Wang}
\author[1]{Yang Chen}
\author[1]{Xiaotong Lin}
\author[1]{Wuliang Huang}
\author[1]{Ziyi Gao}
\author[1]{Xing Fu}
\author[1]{Yu Cheng}
\author[1]{Weiqiang Wang}

\affiliation[1]{DeepFind Team, Ant Group}
\affiliation[2]{East China Normal University}

\contribution[*]{\footnotesize Corresponding author}
\contribution[\dag]{\footnotesize Jiahao Yuan and Yike Xu contributed equally to this work. The work was completed during Jiahao(ECNU)'s internship at Ant Group. \email{jhyuan.cs@gmail.com}}

\abstract{
Decoder-only large language models are increasingly used as behavioral encoders for user representation learning, yet the impact of attention masking on the quality of user embeddings remains underexplored. In this work, we conduct a systematic study of causal, hybrid, and bidirectional attention masks within a unified contrastive learning framework trained on large-scale real-world Alipay data that integrates long-horizon heterogeneous user behaviors. To improve training dynamics when transitioning from causal to bidirectional attention, we propose Gradient-Guided Soft Masking, a gradient-based pre-warmup applied before a linear scheduler that gradually opens future attention during optimization. Evaluated on 9 industrial user cognition benchmarks covering prediction, preference, and marketing sensitivity tasks, our approach consistently yields more stable training and higher-quality bidirectional representations compared with causal, hybrid, and scheduler-only baselines, while remaining compatible with decoder pretraining. Overall, our findings highlight the importance of masking design and training transition in adapting decoder-only LLMs for effective user representation learning. Our code is available at \url{https://github.com/JhCircle/Deepfind-GGSM}.
}
\gtechdata[Correspondence]{\email{yike.wbk@antgroup.com}}

\begin{document}
  \maketitle

\section{Introduction}

User embeddings integrate large-scale heterogeneous signals including textual profiles, interaction histories, and tabular attributes into compact representations that enable robust user understanding in digital marketing \cite{zhang2024scaling}, recommendation \cite{fenglong}, and personalization systems \cite{dou2025transferable,gao2025instruction}. Existing works 
leverage self-supervised contrastive learning to align augmented views of user activity via contextual consistency within sequences \cite{oord2018representation,lin2022dual} or cross-view coherence across sessions \cite{zou2022multi}. Yet as behaviors grow increasingly sequential and context-sensitive, effective representations must anticipate future actions \cite{dou2025transferable}, demanding stronger semantic integration and long-range reasoning.

Bidirectional pre-trained language models (PLMs) address this via full self-attention \cite{devlin2019bert, raffel2020exploring}, enabling holistic embeddings that have dominated general-purpose \cite{wang2024utilizing} and user-centric tasks \cite{sun2019bert4rec,dou2025transferable}. However, their static, batch-oriented design requires the full context upfront—making them impractical for interactive settings where signals arrive incrementally \cite{dou2025transferable}. Decoder-only large language models (LLMs), by contrast, support autoregressive interaction, and recent work shows they can be effective for user modeling when adapted with contrastive objectives \cite{zhang2025qwen3,gao2025instruction}. Crucially, such adaptation hinges on the attention masking strategy: while decoder-only LLMs are pretrained with causal attention, they can be trained and evaluated under three distinct recipes—(i) \textbf{Causal}: standard unidirectional mask \cite{zhang2025qwen3}; (ii) \textbf{Bidirectional}: full self-attention over the entire input \cite{hu2025kalm,li2025conan}; (iii) \textbf{Hybrid}: bidirectional attention over a designated user segment followed by causal attention for downstream tokens. Despite their prevalence, no study systematically compares how these masking choices affect user representation quality under a unified contrastive training framework.

To address this gap, we systematically investigate the role of attention masking and its training dynamics when adapting decoder-only LLMs for user representation learning. Rather than treating masking as a fixed design choice, we highlight the importance of the transition from causal to bidirectional attention and propose a practical warm-up mechanism to stabilize this process within a contrastive learning framework. We evaluate our approach on twelve discriminative user understanding benchmarks derived from Alipay’s real-world user cognition system. Our key contributions are summarized as follows:

\begin{itemize}
  \item We conduct a unified empirical study of \textbf{causal, hybrid, and bidirectional attention masks} for LLM-based user representation learning under a controlled contrastive framework.

  \item We demonstrate that the \textbf{training transition from causal to bidirectional attention} is a key factor affecting optimization stability and representation quality.

  \item We propose \textbf{Gradient-Guided Soft Masking (GG-SM)} as a gradient-informed pre-warmup that facilitates a smoother causal-to-bidirectional transition and leads to stronger final bidirectional representations evaluated on 9 user-centric classification benchmarks.
\end{itemize}

\section{Related Work}
\label{sec:related}
\paragraph{LLM for User Embedding.}Large language models (LLMs) are increasingly employed for user representation learning due to their ability to integrate behavioral sequences, textual profiles, and structured attributes into unified embeddings. Encoder-based models such as BERT4Rec~\cite{sun2019bert4rec} and FOUND~\cite{dou2025transferable} treat user histories as pseudo-sentences and capture rich contextual dependencies, but their bidirectional attention necessitates full input visibility and limits applicability in streaming or interactive scenarios. Decoder-only LLMs overcome this limitation via autoregressive processing, enabling continual updates and dynamic context integration. Recent systems, including Qwen3-embedding~\cite{zhang2025qwen3} and InstructUE~\cite{gao2025instruction}, adapt causal LLMs for embedding tasks through contrastive or instruction-based objectives, yet the impact of attention masking strategies remains underexplored. Existing approaches follow one of three paradigms: causal masking, ensuring compatibility with generative inference; bidirectional masking, maximizing representational completeness but forfeiting autoregressiveness; and hybrid masking, combining bidirectional attention within the history block with causal attention thereafter. Conan~\cite{li2025conan} introduces a progressive scheduler transitioning from causal to bidirectional masking, narrowing the gap between pretraining dynamics and embedding requirements. However, no systematic comparison across these strategies under identical training conditions exists. We fill this gap via large-scale evaluation on 9 real-world user cognition benchmarks, finding that bidirectional masking yields the highest representational quality, while hybrid masking offers the best trade-off between completeness and generative compatibility.

\paragraph{Synthetic Data for User Embedding.}
High-quality labeled data for user modeling remains scarce, motivating growing interest in synthetic data generation. Early methods relied on heuristic augmentation or retrieval-based pseudo-labels~\cite{nogueira2019passage}. Recent approaches leverage large language models to generate realistic behavior traces or user intents~\cite{gao2025instruction}. However, most pipelines depend on proprietary APIs such as GPT-4~\cite{choi2024linq,chen2025little,yuan2025kardia}, which raises concerns about cost, reproducibility, and domain alignment. Alternatives based on small open-source LLMs often suffer from low fidelity due to insufficient semantic alignment with target user behaviors~\cite{wang2024survey}. To improve the quality and scalability of hard positive samples in training data, we propose a training-free synthesis framework that directly leverages an off-the-shelf base LLM to probe hard-to-align user\&query–answer pairs. By applying post-hoc chain-of-thought reasoning, we identify the underlying difficult patterns in these pairs and use these insights to refine prompt for QA synthesis, enabling scalable generation of high-fidelity synthetic hard positives.
\section{Training Data}
\label{sec:training}
Following \cite{dou2025transferable,gao2025instruction}, we contrust and employ two types of alignment data for embedding training based on real-world Alipay user interactions:
\textbf{(1) Rule-based Behavioral Trajectories Dataset $ \mathcal{D}_{behavior}$:} Composed of user behavior sequences, $ \mathcal{D}_{behavior} = \{ u_i, b_i \}_{i=1}^{N} $, where $b_i$ denotes the user $i$'s real future behavior.  
\textbf{(2) LLM-synthesized Query–Answer Alignments Dataset $ \mathcal{D}_{qa}$:} Represents user intent and language understanding, denoted as $ \mathcal{D}_{qa} = \{ u_i \oplus q_i, a_i \} $, where $q_i$ is the query generated from $u_i$, and $a_i$ is the corresponding LLM-generated answer. Here, $\mathbf{u}_i = \{ Bill_i, Mini_i, Spm_i, App_i, Search_i, Tabular_i \} \in \mathcal{U}$ represent the user $i$’s multi-modal interaction profile over the past 90 days, where $Bill_i$ denotes PayBill transactions, $Mini_i$ represents Mini Program interactions, $Spm_i$ represents super position model (SPM) paths $S_i$ refers to superposition model paths, $S_i$ captures search queries, and $T_i \in \mathbb{R}^{F \times D}$ includes tabular features with $F$ features and $D$-dimensional embeddings.

\subsection{Rule-based Behavioral Trajectories Dataset}
\label{subsec:rule}
Following \cite{dou2025transferable}, we construct behavioral trajectory pairs using a rule-based alignment strategy. The left tower encodes the user's raw interaction sequence over the past three months. For the right tower, we first aggregate all interactions from the subsequent one-month window (e.g., by action type or temporal bins), then randomly sample a representative subset to serve as the future prediction target. This aggregated-and-sampled future signal is aligned with the historical sequence during embedding training.

\subsection{LLM-Synthesized Query-Answer Alignments Dataset}
\label{subsec:llmdata}

Building on the insights from \cite{robinsoncontrastive,lee2024difficulty} that challenging negative samples can enhance embedding learning, we extend this idea by focusing on generating challenging positive samples as anchors through a post-rule improvement mechanism, optimizing data synthesis by pre-generating challenging query-answer pairs to avoid the embedding-based real-time post-mining of negative samples constrained by data quality \cite{gao2025instruction} and computational limitations \cite{li2025conan} during training.
\paragraph{Step (1) Synthesis Pipeline and Calibration Set Generation.}
We begin by initializing our synthesis pipeline with Qwen-Max to generate diverse user-understanding scenarios as a seed pool $Pool$ for the subsequent synthesis of varied and generalizable QA pairs. Given each user $i$, we prompt LLM $\mathcal{LLM}$ to retrieve the top 10 most relevant seed scenario $seed_{top10}$ according to user behavior history $u_i$ via $P_{retrieve}$ and then instantiate $u_i$ and $seed_{top10}$ to synthesize QA pairs that reflect diverse user understanding through prompting $\mathcal{LLM}$ with $P_{qa}$. From these, we construct a \textbf{calibration set} $\mathcal{D}_{c}$ of 1,000 user\&query-answer pairs $\{(u_i \oplus q_i, a_i)\}_{i=1}^{1000}$, ensuring diverse coverage of user behavior topics, formally:
\begin{align}
\label{eq:syn}
seed_{\text{top10}}(u_i, \mathcal{P}) &= \mathcal{LLM}(u_i, Pool, P_{retrieve}), \\
\mathcal{D}_c &= \{\mathcal{LLM}(seed_{\text{top10}}, u_i, P_{qa})\}_{j=1}^{1000}.
\end{align}
\paragraph{Step (2) Alignment Difficulty Probing on $\mathcal{D}_c$.}
Inspired by difficulty probing in \cite{team2024qwen2}, for each user\&query-answer pair $(u_i \oplus q_i, a_i)$, we evaluate its alignment difficulty by computing the similarity between $u_i \oplus q_i$ and $a_i$ as hard-to-align score $S_{d}$ via a strong and size-efficient embedding model $Emb$ \footnote{\url{https://huggingface.co/Qwen/Qwen3-Embedding-0.6B}}, formally: 
\begin{equation}
    S_{d}= 1-\texttt{Sim}(Emb(u_i \oplus q_i), Emb(a_i))
\end{equation}
where $\texttt{Sim}(v_1, v_2)$ represents the cosine similarity between $v_1$ and $v_2$, computed as: $\texttt{Sim}(v_1, v_2) = \frac{v_1 \cdot v_2}{\|v_1\| \|v_2\|}$. Higher $S_d$ values indicate more difficult alignments, meaning the pair $(u_i \oplus q_i, a_i)$ is harder to align even for a strong original model. We further set a threshold $T_{filter}$ to retain only the \textbf{challenging samples} $\mathcal{D}_{hard} = \{ (u_i \oplus q_i, a_i) \mid S_d \geq T_{filter}\} $ with high alignment difficulty.

\paragraph{Step (3) Inductive Feature Completion.}
Inspired by post-cot interpretation for feature interpretation \cite{singh2024rethinking}, for the remaining challenging pairs, we apply an inductive \textbf{feature completion rule} using \textbf{Qwen-Max} to extract common rules $P_{\text{rule}}$ from hard-to-align positive QA pairs $\mathcal{D}_{hard}$ and further integrate them into prompt $P_{qa}$ for qa synthesis as described in Eq.~\ref{eq:syn}.

\paragraph{Step (4) Scaling and Posterior Rewriting.}
After enriching the challenging query-answer pairs, we scale the step (1) by applying the optimized $P_{qa}$ to generate a larger set of query-answer pairs following the input template specified in Appendix~\ref{app:training}, which standardizes modality delimiters, instruction formatting, and the placement of the special \texttt{<USER>} token. These pairs undergo posterior rewriting to ensure enhanced clarity, context alignment, and semantic consistency with the user's historical behavior. This step refines the dataset, preparing the pairs for embedding model training and improving their alignment with real-world user interactions.

\begin{figure*}[t] 
\centering
\includegraphics[width=\textwidth]{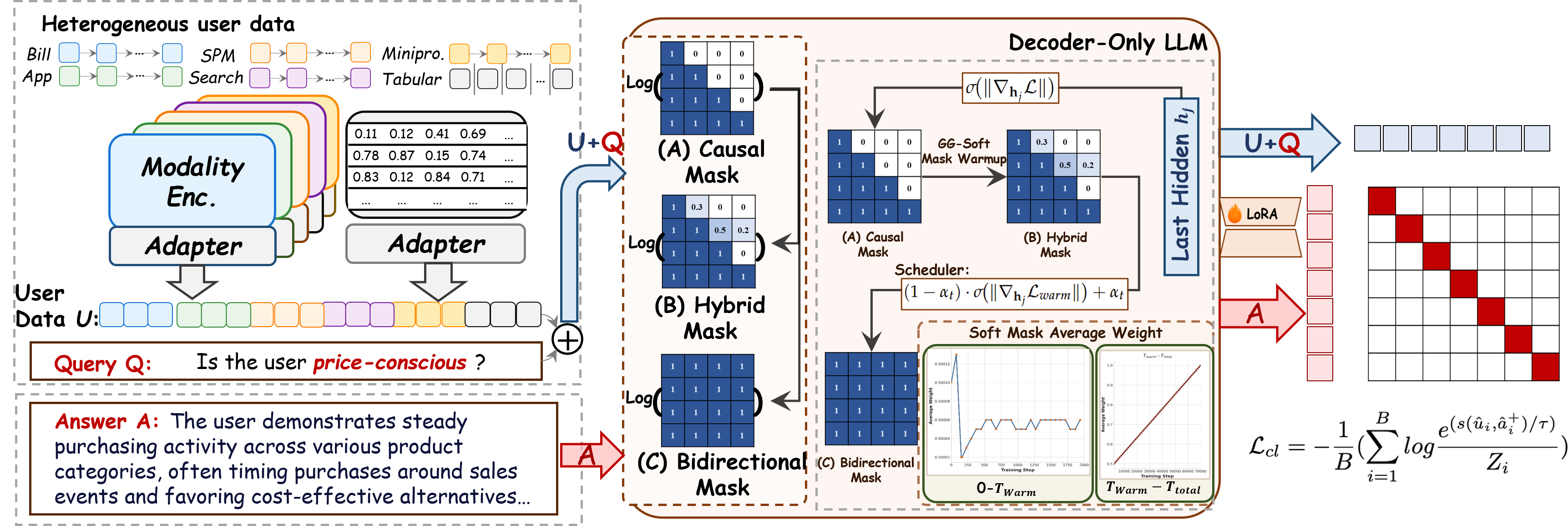} 
\caption{Architechure Overview of Our Find-Embedding (w / GGSM).} 
\label{Fig.frame} 
\end{figure*}

\section{LLMs as Encoders: Training Recipe}
\label{sec:recipe}
\paragraph{Training Architecture.}As illustrated in Figure~\ref{Fig.frame}, our framework follows \cite{dou2025transferable} to process user pair $u_i \oplus q_i$ via modality-specific encoders whose outputs are projected into the LLM’s ($\mathcal{M}$) embedding space via lightweight adapters. In parallel, the corresponding answer $a_i$ through the same decoder-only LLM $\mathcal{M}$ as a dual-tower alignment architecture. Both towers share the LLM backbone but operate independently during encoding, enabling efficient, modality-aware representation learning while maintaining compatibility with the LLM’s token semantics for downstream contrastive alignment. Implement details are provided in Sec.~\ref{sec:exp} and representation learning procedures are deferred to Appendix~\ref{app:training}.

\paragraph{Gradient-Guided Soft Masking.}  
To endow causal LLMs with bidirectional reasoning capabilities during encoding, we extend the causal-to-bidirectional scheduler of \citet{li2025conan} with a gradient-guided warmup phase. Let $T_{warm}$ and $T_{total}$ denote the warmup and total training steps, respectively. For a sequence of user\&query length $L$ with hidden states $\mathbf{H} = [\mathbf{h}_1, \dots, \mathbf{h}_L] \in \mathbb{R}^{L \times d}$, we define the soft attention mask $M^{\text{soft}}(t) \in \mathbb{R}^{L \times L}$ at training step $t$ as:
\begin{align}
M^{\text{soft}}_{ij}(t) &=
\begin{cases}
0 & \text{if } j \leq i, \\
\log w_{ij}(t) & \text{if } j > i,
\end{cases} \\
w_{ij}(t) &=
\begin{cases}
\sigma\!\big(\|\nabla_{\mathbf{h}_j} \mathcal{L}\|\big) & \text{if } t < T_{warm}, \\
(1 - \alpha_t) \cdot \sigma\!\big(\|\nabla_{\mathbf{h}_j} \mathcal{L}_{warm}\|\big) + \alpha_t & \text{if } T_{warm} \leq t < T_{total}.
\end{cases}
\end{align}
where $\alpha_t = \frac{t - T_{warm}}{T_{total} - T_{warm}} \in [0,1]$ and $\mathcal{L}_{warm}$ denotes the loss computed at the final warmup step $t = T_{warm} - 1$, and $\sigma(\cdot)$ is the sigmoid function ensuring $w_{ij}(t) \in (0,1]$. During warmup ($t < T_{warm}$), future attention weights are set adaptively via the instantaneous gradient norm $\|\nabla_{\mathbf{h}_j} \mathcal{L}\|$: tokens that strongly influence the loss receive higher visibility. At the end of warmup, these gradient-derived weights are frozen. In the scheduler phase ($t \geq T_{warm}$), we linearly interpolate between the frozen soft mask and full bidirectionality (i.e., $w_{ij} = 1$), and at inference, $\mathcal{M}$ employs a fully bidirectional attention mask to maximize contextual integration, thereby better bridge the gap between token-level pretraining with sentence-level representation learning \cite{li2025conan}.

\paragraph{Training Objective.}

We employ contrastive learning, following \cite{li2023towards,zhang2025qwen3}, to align user and answer embeddings. The goal is to learn discriminative user embeddings by pulling semantically related user-answer pairs closer and pushing apart negative samples, facilitating comprehensive user profile extraction and accurate future action prediction. The objective is defined using the InfoNCE loss $\mathcal{L}_{cl}$ across a batch of size $B$, formally:

\begin{equation}
\mathcal{L}_{cl} = -\frac{1}{B} \sum_{i=1}^{B} \log \frac{e^{s(\hat{u}_i, \hat{a}_i^+) / \tau}}{Z_i},
\end{equation}

where $\hat{u}_i$ and $\hat{a}_i$ are the normalized embeddings of user $i$ and its answer. $s(\hat{u}_i, \hat{a}_i)$ is the cosine similarity between the user and answer embeddings, and $\tau$ controls the similarity smoothness. $Z_i$ is the normalization factor, aggregating positive and negative pair similarities:
\begin{align}
Z_i = & e^{s(\hat{u}_i, \hat{a}_i^+) / \tau} + \sum_{j \neq i} m_{ij} e^{s(\hat{u}_i, \hat{a}_j) / \tau} \nonumber \\
& + \sum_{j \neq i} m_{ij} e^{s(\hat{u}_i, \hat{u}_j) / \tau} + \sum_{j \neq i} m_{ij} e^{s(\hat{a}_i, \hat{a}_j) / \tau},
\end{align}
where $a_i^+$ and $a_j$ are the positive and other in-batch answer embeddings, respectively, and $u_j$ represents other in-batch user embeddings. To mitigate the effect of false negatives, we follow \cite{zhang2025qwen3} and introduce a mask factor $m_{ij}$, which ensures that negative samples are sufficiently distinct. The mask factor is computed as:
\begin{equation}
m_{ij} =
\begin{cases}
    0 & \text{if } s_{ij} > s(\hat{u}_i, \hat{a}_i^+) + c_{\text{margin}}, \\
    1 & \text{otherwise}.
\end{cases}
\end{equation}
where $s_{ij}$ represents the similarity between user embeddings $\hat{u}_i, \hat{u}_j$ or mixed embeddings $\hat{u}_i, \hat{a}_j$, and $c_{\text{margin}}$ is a margin hyperparameter that ensures adequate separation between positive and negative pairs. Incorporating same-side negative samples enhances the distinctiveness of user representations across different instructions and improves the separability of answer embeddings, ultimately boosting model performance in embedding-based tasks.

\section{Experiments}
\label{sec:exp}
\paragraph{Models and Implementation.} For training data, we follow our pipeline (Sec.~\ref{subsec:llmdata}) via \texttt{Qwen3-30B-A3B} \footnote{\url{https://huggingface.co/Qwen/Qwen3-30B-A3B}} \cite{qwen3technicalreport} for efficiency. For training 
architecture, we adopt dedicated instances of \texttt{gte-base-zh} \cite{li2023towards} to encode heterogeneous behavioral inputs into modality-specific embeddings, which are concatenated and prepended to the input of \texttt{Qwen2.5-0.5B-Instruct} \cite{team2024qwen2}, serving as the LLM backbone for contrastive user representation learning. And we fine-tune this decoder-only LLM under a contrastive learning objective with distinct attention masking strategies (detailed in Sec.~\ref{sec:recipe}), using identical training configurations across all variants: a global batch size of 2,048, 7w fine-tuning steps, an AdamW optimizer with initial learning rate $2 \times 10^{-4}$ and cosine decay, LoRA \cite{hulora} with rank=64 and $\alpha$=64. All experiments are trained on 64 A100-80GB GPUs using data parallelism, while inference is performed on single A100-80GB GPU for subsequent evaluation.

\paragraph{Baselines and Tasks.} We select Qwen2.5-0.5B-Instruct as the oracle backbone and compare its performance under three attention mask training recipes: (1) Causal: contrastive learning with the original causal attention mask; (2) Hybrid: three strategies for opening the upper triangular attention matrix: (a) gradient-guided soft masking using left tower gradients to compute importance scores that control the future mask, (b) applying an MLP for direct attention opening, and (c) introducing a global query in a CLS-like fashion to guide attention; (3) Bidirectional: (a) contrastive learning with the bidirectional mask, and transitioning from unidirectional to bidirectional training via (b) scheduler or \textbf{(c) gradient-guided soft mask pre-warmup and scheduling (Ours)}. All recipes are detailed in Appendix~\ref{app:detail}. Additionally, we evaluate inference performance using the top three embedding models \footnote{Valid during the working period until December 31, 2025.} from the MTEB leaderboard including KaLM-Embedding-Gemma3-12B-2511 \cite{hu2025kalm}, llama-embed-nemotron-8b \cite{babakhin2025llama}, Qwen3-Embedding-8B \cite{zhang2025qwen3}. For broader comparison, we also include representative traditional user modeling baselines, where U-MLP One4all~\cite{shin2021one4all} extends a general-purpose One4all representation with an additional MLP decoder for user targeting, while MSDP~\cite{fu2023robust} and CPC~\cite{oord2018representation} adopt contrastive learning to learn robust user representations from augmented views of behavior sequences, together with LLM-based user representation models such as FOUND~\cite{dou2025transferable}. All models are evaluated under consistent training hyperparameters, with the evaluation performed on a binary classification task across 9 real-world Alipay user scenarios, as listed in Table~\ref{tab:task}:

\renewcommand{\arraystretch}{1.1}
\begin{table}[H]
\centering
    \begin{tabular*}{0.95\linewidth}{@{\extracolsep{\fill}} p{0.05\linewidth} p{0.25\linewidth} p{0.4\linewidth} p{0.15\linewidth} }
    \hline
         Dataset & Domain & Scenario & Number \\
         \hline
         $\mathcal{D}_{train}$ 
         & General~(\ref{subsec:rule}~\&~\ref{subsec:llmdata}) 
         & General 
         & $\approx$ $1.433 \times 10^8$ \\
         \hline
         \multirow{4}{*}{$\mathcal{D}_{test}$} 
         & \ding{182} User Prediction 
         & Concert Click Prediction (Concert), User Log-in Prediction (User), MAU Loss Prediction (MAU)
         & $\approx$ 50w per task \\
         & \ding{183} Behavior Preference 
        & Public Transit Preference (Transit), Consumption Power (Power), Food Interest (Food), Movie Interest (Movie)
         & $\approx$ 50w per task \\
         & \ding{184} Marketing Sensitivity 
         & Achievement Preference (Achiev.), Physical Preference (Physical)
         & $\approx$ 50w per task \\
         \hline
    \end{tabular*}
    \caption{Data information for user pretraining and test benchmarks, with number of tests per task.}
    \label{tab:task}
\end{table}

\paragraph{Evaluation Metrics.} We assess user representations via linear probing on 9 annotated binary classification tasks from Alipay’s user cognition system, reporting AUC (Area Under the ROC Curve~\cite{bradley1997use}) for discriminative performance.

\section{Main Results}

\begin{figure*}[ht] 
\centering
\includegraphics[width=\textwidth]{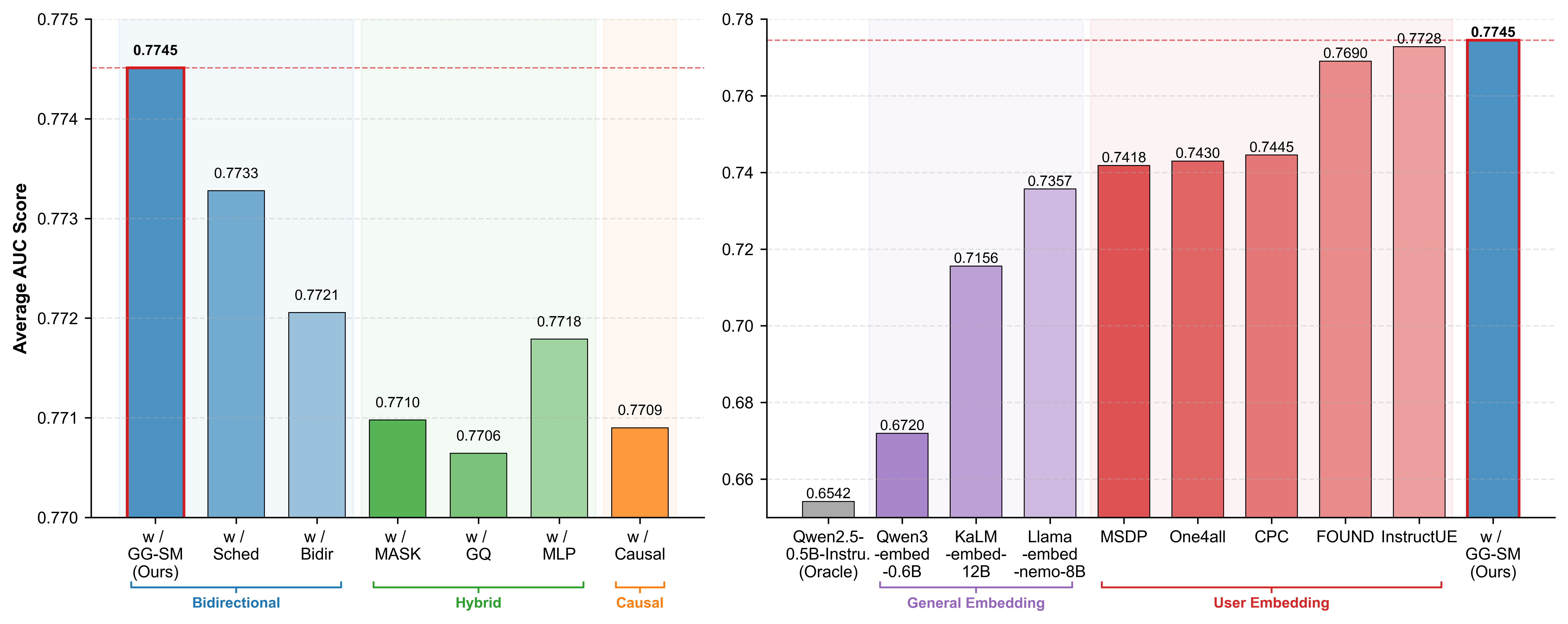} 
\caption{
Average AUC performance across 9 downstream tasks under different attention masking strategies (left) and comparison with general embedding, user embedding (right).
}

\label{fig:main_results} 
\end{figure*}

\begin{table}[!h]
\renewcommand{\arraystretch}{1.2}
\centering
\footnotesize
\begin{adjustbox}{width=\linewidth}
\begin{tabular}{lccccccccccc}
\toprule
& \multicolumn{3}{c}{\textbf{User Prediction}} 
& \multicolumn{4}{c}{\textbf{Behavior Preference}} 
& \multicolumn{2}{c}{\textbf{Marketing Sensitivity}} \\
\cmidrule(lr){2-4} \cmidrule(lr){5-8} \cmidrule(lr){9-10}

\textbf{Method} 
& \textbf{Concert} 
& \textbf{User} 
& \textbf{MAU} 
& \textbf{Transit} 
& \textbf{Power} 
& \textbf{Food} 
& \textbf{Movie} 
& \textbf{Achiev.} 
& \textbf{Physical} 
& \textbf{Avg} \\
\midrule

\multicolumn{11}{l}{\textit{\textbf{General Embedding Models}}} \\
Qwen3-Embedding-0.6B     
& 0.5226
& 0.7294
& 0.9197
& 0.6098
& 0.8078
& 0.6656
& 0.6641
& 0.5529
& 0.5759
& 0.6720 \\

Llama-embed-nemotron     
& 0.5627
& 0.7735
& 0.9351
& 0.6915
& 0.9308
& 0.7936
& 0.7692
& 0.5879
& 0.5768
& 0.7357 \\

KaLM-Embedding           
& 0.5359
& 0.7609
& 0.9272
& 0.6400
& 0.8623
& 0.7443
& 0.7812
& 0.5787
& 0.6099
& 0.7156 \\
\midrule

\multicolumn{11}{l}{\textit{\textbf{User Embedding Models}}} \\
MSDP \cite{fu2023robust} 
& 0.5155
& 0.9504
& 0.9633
& 0.6367
& 0.8480
& 0.7928
& 0.7645
& \underline{0.6151}
& 0.5900
& 0.7418 \\

One4all \cite{shin2021one4all} 
& 0.5568
& \textbf{0.9509}
& 0.9639
& 0.6276
& 0.8393
& 0.7984
& 0.7526
& 0.6016
& 0.5957
& 0.7430 \\

CPC \cite{oord2018representation} 
& 0.5314
& \underline{0.9506}
& 0.9654
& 0.6376
& 0.8415
& 0.8009
& 0.7526
& \textbf{0.6256}
& 0.5952
& 0.7445 \\

FOUND \cite{dou2025transferable} 
& 0.5670
& 0.8330
& 0.9574
& 0.6824
& 0.9669
& 0.8513
& 0.8472
& 0.6102
& 0.6059
& 0.7690 \\

InstructUE \cite{gao2025instruction} 
& 0.5712
& 0.8394
& 0.9661
& 0.6964
& \textbf{0.9695}
& 0.8534
& \textbf{0.7927}
& 0.6071
& 0.6594
& 0.7728\\
\midrule

\multicolumn{11}{l}{\textit{\textbf{Qwen2.5-0.5B-Instruct (Causal)}}} \\
Oracle                   
& 0.5173
& 0.7219
& 0.9202
& 0.5642
& 0.7638
& 0.6561
& 0.6435
& 0.5415
& 0.5592
& 0.6542 \\

w/ Causal                 
& 0.5716
& 0.8313
& 0.9669
& 0.6967
& 0.9678
& 0.8473
& 0.7922
& 0.6054
& 0.6589
& 0.7709 \\
\midrule

\multicolumn{11}{l}{\textit{\textbf{Qwen2.5-0.5B-Instruct (Hybrid)}}} \\
w/ $Hybrid_{\text{mask}}$   
& 0.5748
& 0.8311
& 0.9671
& 0.6951
& 0.9653
& 0.8520
& 0.7913
& 0.6056
& 0.6565
& 0.7710 \\

w/ $Hybrid_{\text{gq}}$     
& 0.5647
& 0.8382
& 0.9665
& 0.6945
& 0.9678
& 0.8528
& 0.7887
& 0.6044
& 0.6582
& 0.7706 \\

w/ $Hybrid_{\text{mlp}}$    
& \underline{0.5750}
& 0.8410
& 0.9667
& 0.6965
& 0.9649
& 0.8484
& 0.7886
& 0.6042
& 0.6608
& 0.7718 \\
\midrule

\multicolumn{11}{l}{\textit{\textbf{Qwen2.5-0.5B-Instruct (Bidirectional)}}} \\
w/ Bidirectional         
& 0.5707
& 0.8390
& 0.9673
& \textbf{0.6983}
& 0.9671
& 0.8505
& 0.7906
& 0.6043
& \underline{0.6607}
& 0.7721 \\

w/ Scheduler               
& 0.5742
& 0.8419
& \underline{0.9664}
& 0.6973
& 0.9688
& \underline{0.8540}
& 0.7908
& 0.6056
& 0.6605
& \underline{0.7733} \\

\rowcolor[RGB]{236,244,252}
\textbf{w/ GG-SM (Ours)}   
& \textbf{0.5767}
& 0.8438
& \textbf{0.9674}
& \underline{0.6978}
& \underline{0.9689}
& \textbf{0.8554}
& \underline{0.7913}
& 0.6078
& \textbf{0.6615}
& \textbf{0.7745} \\
\bottomrule
\end{tabular}
\end{adjustbox}

\caption{Comparison of AUC performance for general embeddings, user embeddings, and our GG-SM method across all downstream tasks.}
\label{tab:tab_results}
\end{table}

In this section,  we evaluate the effectiveness of our proposed \textbf{GG-SM} training strategy (Sec.~\ref{sec:recipe}) across 9 downstream user-centric tasks. Figure~\ref{fig:main_results} illustrates the average AUC across tasks under different attention masking strategies (left) and a comparison with general embeddings, user embeddings, Oracle, and Ours (right). Table~\ref{tab:tab_results} provides detailed numerical results across three major domains: \textit{User Prediction}, \textit{Behavior Preference}, and \textit{Marketing Sensitivity}.

\paragraph{Parameter Efficiency and Domain-Specific Alignment.} 
A primary finding from Table~\ref{tab:tab_results} is that our GG-SM-enhanced Qwen2.5-0.5B-instruct achieves an average AUC of \textbf{0.7745}, consistently outperforming massive general-purpose embeddings such as \textbf{Llama-embed-nemotron} (0.7357) and \textbf{KaLM-Embedding} (0.7156). With significantly fewer parameters, GG-SM still outperforms on task-specific metrics (\textit{Transit}: 0.6978; \textit{Power}: 0.9689), verifying that raw parameter scale yields diminishing returns when applied to industrial behavioral logs with high sparsity and non-linguistic distributions. While 8B+ models possess broader natural language priors, they introduce redundant noise in discrete behavioral sequences. In contrast, GG-SM maximizes \textit{information extraction density}, proving that gradient-based attention calibration is more critical than raw scaling for aligning an LLM's latent space with domain-specific behavioral structures.

\paragraph{From Local Contrast to Contextual Priors.} 
The results highlight a decisive performance gap between traditional user modeling and LLM-based approaches. Traditional baselines like \textbf{MSDP}, \textbf{One4all}, and \textbf{CPC} excel in specific tasks like \textit{Achiev.} (\textbf{0.6256}), likely due to their effective capture of local feature matches. However, they struggle with tasks requiring \textbf{global contextual transfer}, such as \textit{Food} or \textit{Movie} preferences. Comparing our model against recent LLM-based baselines like \textbf{FOUND} (0.7690) and \textbf{InstructUE} (0.7728), we observe that GG-SM provides more consistent gains. While these models utilize LLMs as static extractors, GG-SM treats the attention mechanism as an evolvable bottleneck, effectively leveraging pre-trained contextual priors to model long-range user dependencies more holistically.

\begin{wrapfigure}{r}{0.4\textwidth}
\centering
\vspace{-3pt}
\includegraphics[width=0.4\textwidth]{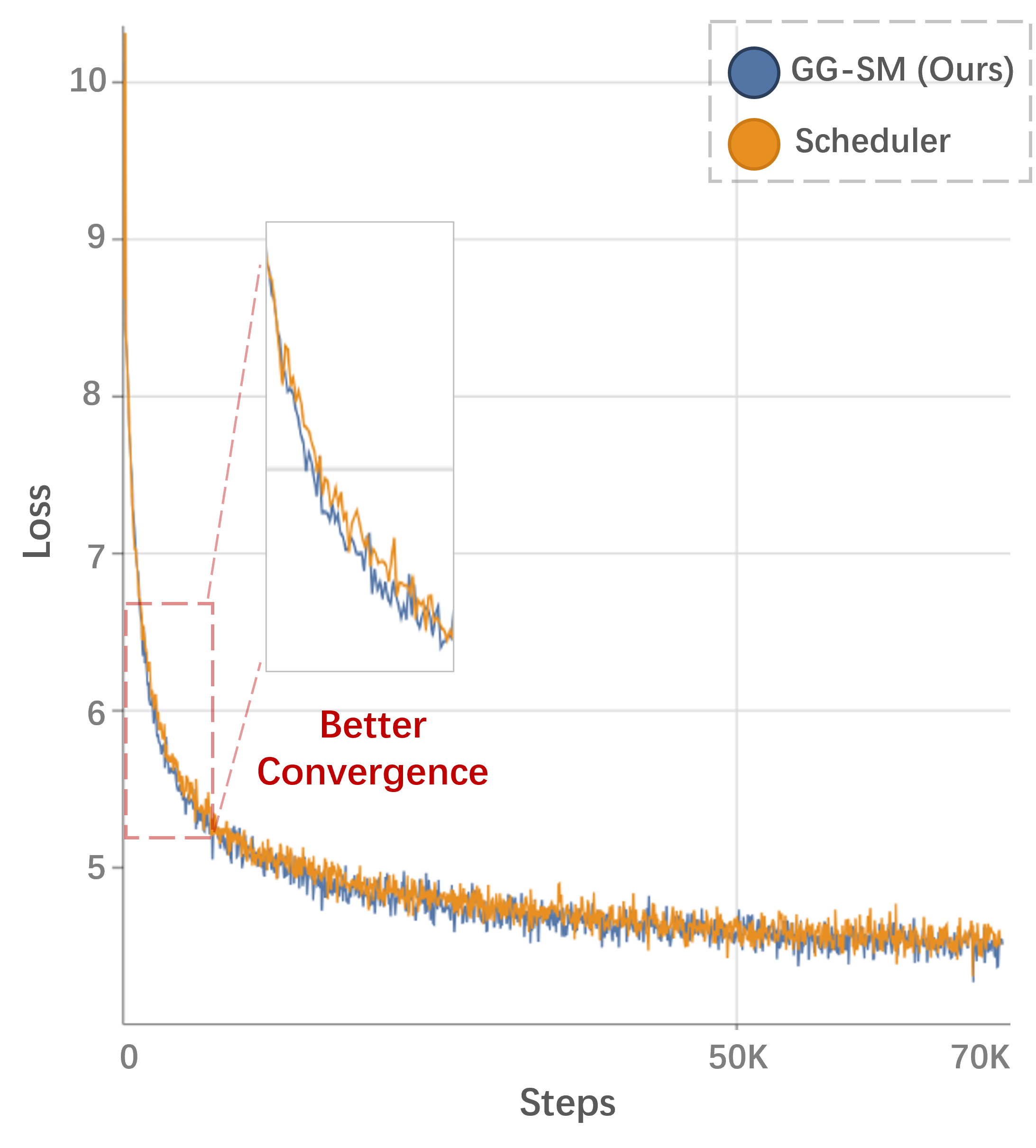}
\caption{Training loss convergence: GG-SM (Ours) vs. scheduler.}
\label{fig:loss_convergence}
\end{wrapfigure}
\paragraph{Efficacy of Gradient-Guided Attention Evolution.} 
The internal comparison of masking strategies reveals that the \textit{path} to bidirectionality determines the quality of the final embedding. Standard Causal masks (Oracle) are too restrictive for representation tasks, while Hybrid strategies ($Hybrid_{\text{mask}}$, $Hybrid_{\text{gq}}$, $Hybrid_{\text{mlp}}$) provide only marginal gains as they introduce additional parameters that are difficult to align with a frozen or pre-trained backbone. GG-SM outperforms both naive Bidirectional and Scheduler-based methods by using \textbf{instantaneous gradient norms} as a dynamic signal for token importance. This ensures that the model does not merely see more tokens, but learns to prioritize the most informative ones during the crucial early stages of bidirectional adaptation as shown in Fig.~\ref{fig:loss_convergence}. As evidenced by the \textit{Behavior Preference} results, this leads to a significantly sharper separation of user interests compared to static masking recipes.

\paragraph{Domain Robustness and Transferability.} 
The robustness of GG-SM is evidenced by its consistent lead across three distinct domains: \textit{User Prediction}, \textit{Behavior Preference}, and \textit{Marketing Sensitivity}. While traditional contrastive models often exhibit high variance—e.g., performing well in high-frequency behavior tasks but degrading in sensitivity tasks—our model maintains a stable performance advantage. Notably, in the \textbf{Marketing Sensitivity} domain, where latent intent is hardest to capture, GG-SM achieves peak AUC. This suggests that guiding the attention mechanism through the model's own internal learning pressure (gradients) captures more transferable user traits than manually engineered data augmentations or fixed architectural modifications.
\section{Conclusion}

In this work, we revisit user representation learning with decoder-only LLMs through the lens of attention masking, systematically comparing causal, hybrid, and bidirectional masks under a unified contrastive framework on large-scale real-world Alipay data and 9 industrial user-centric tasks. We show that not only the final mask but also the transition path from causal to bidirectional modeling critically affects training stability and embedding quality. To this end, we introduce Gradient-Guided Soft Masking as a pre-warmup before a linear scheduler, which consistently improves optimization behavior and yields stronger bidirectional representations while remaining compatible with decoder pretraining. Overall, our findings highlight that careful masking design and transition dynamics are key to effectively adapting decoder-only LLMs as practical user encoders.
\bibliographystyle{assets/plainnat}
\bibliography{main}

\appendix
\section{Details of Representation Training}
\label{app:training}
\subsection{Standardized Input Template}
\begin{figure}[ht]
\begin{tcolorbox}[
    width=\linewidth,
    colback=white,
    colframe=black,
    sharp corners,
    boxrule=1pt,
    top=6pt,
    bottom=6pt
]
The following presents heterogeneous user data collected from multiple sources, including PayBill transactions ($B_i$), Mini Program interaction logs ($M_i$), Super Position Model paths ($S_i$), App interaction records ($A_i$), homepage search queries ($\mathcal{R}_i$), and structured tabular features ($T_i$):
\texttt{<bill>} \{ Bill data $B_i$ \} \texttt{</bill>} \\
\texttt{<minipro>} \{ Mini Program logs $M_i$ \} \texttt{</minipro>} \\
\texttt{<spm>} \{ Super Position Model paths $S_i$ \} \texttt{</spm>} \\
\texttt{<app>} \{ App interaction records $A_i$ \} \texttt{</app>} \\
\texttt{<search>} \{ Search queries $\mathcal{R}_i$ \} \texttt{</search>} \\
\texttt{<tabular>} \{ Tabular features $T_i \in \mathbb{R}^{F \times D}$ \} 
\texttt{</tabular>} \\
\textbf{Instruction:} \{ optional user query $q_i$ \} \ \textcolor{red}{\texttt{<USER>}}
\end{tcolorbox}
\caption{Input format of Training Data. \textcolor{red}{\texttt{<USER>}} token serves as the anchor for extracting user embedding.
}
\label{fig:template}
\end{figure}
To ensure consistency across modalities and reproducibility of representation learning, 
we adopt a unified input template for all synthesized and real-world alignment data. 
As illustrated in Figure~\ref{fig:template}, each user instance $u_i$ is represented as a heterogeneous sequence of multimodal records collected over the past 90 days, formally:

\begin{equation}
\mathbf{u}_i = \{ Bill_i, Mini_i, Spm_i, App_i, Search_i, T_i \} \in \mathcal{U},
\end{equation}

where each modality is enclosed by explicit semantic boundary tokens, e.g., 
\texttt{<bill>}...\texttt{</bill>}, \texttt{<minipro>}...\texttt{</minipro>}, etc., 
to preserve modality structure and facilitate modality-aware encoding.

For each user instance, a (optional) user instruction $q_i$ may be appended after the user profile. 
The complete model input is formulated as:

\begin{equation}
x_i = u_i \oplus [q_i] \oplus \texttt{<USER>},
\end{equation}

where $\oplus$ denotes sequence concatenation, and $[q_i]$ indicates that the instruction is optional. 
The special token \texttt{<USER>} signals the model to aggregate all preceding multimodal information 
(and the instruction, if provided) into a unified user representation.

\subsection{User Embedding Extraction}

We adopt a decoder-only causal LLM $\mathcal{M}$ as the backbone encoder. 
Given an input sequence $x_i$ of length $L$, the model produces a sequence of final-layer hidden states:

\begin{equation}
\mathbf{H}_i = [\mathbf{h}_1, \dots, \mathbf{h}_L] = \mathcal{M}(x_i),
\end{equation}

where $\mathbf{h}_t \in \mathbb{R}^d$ denotes the hidden state of the $t$-th token.

Let $t_{\text{user}}$ denote the position of the special token \texttt{<USER>} in $x_i$. 
The unified user embedding is defined as the corresponding hidden state:

\begin{equation}
\hat{u}_i = \mathbf{h}_{t_{\text{user}}}.
\end{equation}

To ensure stable contrastive training, we apply $L_2$ normalization:

\begin{equation}
\tilde{u}_i = \frac{\hat{u}_i}{\|\hat{u}_i\|_2}.
\end{equation}

\subsection{Answer Embedding Extraction}

For each answer $a_i$, we feed it independently into the same LLM backbone $\mathcal{M}$ and append an end-of-sequence token \texttt{<EOS>}. The answer embedding is extracted analogously:

\begin{equation}
\hat{a}_i = \mathbf{h}_{t_{\text{EOS}}}, \quad
\tilde{a}_i = \frac{\hat{a}_i}{\|\hat{a}_i\|_2}.
\end{equation}

These normalized embeddings $\tilde{u}_i$ and $\tilde{a}_i$ are then used for contrastive alignment via the InfoNCE objective in Sec.~\ref{sec:recipe}.
\subsection{Training Recipes Across Causal, Hybrid, and Bidirectional Masking}
\label{app:detail}
We release the complete implementation details for all baselines to ensure transparency and reproducibility in this technique report, with particular emphasis on our hybrid masking variants.  We explicitly frame the hybrid approach as \textit{user-centric}, designed to better capture contextual directionality in user representation learning. We regard hybrid masking—especially the progressive hybrid‑to‑bidirectional transition—as a promising research direction, and we encourage the community to further explore and advance this paradigm. Below, w  e outline the exact definitions:

\paragraph{Causal Masking.}
The causal masking strategy employs standard autoregressive attention, where each token $t_i$ attends only to tokens $\{t_j \mid j \leq i\}$. Formally, the attention mask $M^{\text{causal}} \in \mathbb{R}^{L \times L}$ for a sequence of length $L$ is defined as:
\[
M^{\text{causal}}_{ij} = 
\begin{cases}
0 & \text{if } j \leq i, \\
-\infty & \text{otherwise}.
\end{cases}
\]
This enforces strict left-to-right information flow, preserving compatibility with generative inference and the pretraining dynamics of decoder-only LLMs. We apply the contrastive learning objective directly on representations extracted from this causal encoder.

\paragraph{Hybrid Masking.}
Hybrid masking selectively relaxes causality over the user-history segment while maintaining causal constraints for future tokens. We implement three user-centric variants:  

\begin{enumerate}
\item[(a)] \textit{Gradient-Guided Soft Masking}: During training, we compute importance scores for future positions using gradients from a frozen left-tower encoder as illustrated in Sec.~\ref{sec:recipe}. 
\item[(b)] \textit{MLP-Driven Attention Opening}: A lightweight MLP predicts attention bias for $j > i$, dynamically enabling direct future token access based on $\mathbf{h}_i$.  
 \item[(c)] \textit{Global-Query Guidance}: A learnable [CLS]-like token $\mathbf{q}_{\text{global}}$ attends bidirectionally to all history tokens; its attention weights supervise block-level contextual integration without violating causality for downstream generation.
\end{enumerate}

\paragraph{Bidirectional Masking.}
Bidirectional masking grants full self-attention (all-to-all token visibility) and is instantiated in three ways:  
\begin{enumerate}
\item[(a)] \textit{Direct Bidirectional Contrastive}: The model uses a fully unmasked attention matrix $M^{\text{bi}}_{ij} = 0$ for all $i,j$ from initialization, trained with the same contrastive objective as other variants.  
\item[(b)] \textit{Scheduler-Based Transition}: The attention span grows from causal to bidirectional via a deterministic schedule; e.g., at epoch $e$, the mask allows attention up to position $\min(i + \Delta(e), L)$, where $\Delta(e)$ increases linearly or cosinely with $e$.  
\item[(c)] \textit{Gradient-Guided Soft-Mask Warm-Up (Ours)}: Building on the hybrid approach, we first warm up with gradient-derived soft masks (as in Hybrid (a)), then linearly interpolate toward full bidirectionality. Specifically, for step $t \geq T_{\text{warm}}$, the future mask weight is:
\begin{equation}
w_{ij}(t) = (1 - \alpha_t) \cdot \sigma\!\big( \|\nabla_{\mathbf{h}_j} \mathcal{L}_{\text{warm}}\| \big) + \alpha_t, \quad \alpha_t = \frac{t - T_{\text{warm}}}{T_{\text{total}} - T_{\text{warm}}}
\end{equation}

\end{enumerate}
where $\mathcal{L}_{\text{warm}}$ is the loss at the end of warm-up. This data-driven transition enables stable convergence to a fully bidirectional encoder while leveraging task-specific signal during adaptation.

\end{document}